# Unity RL Playground: A Versatile Reinforcement Learning Framework for Mobile Robots

Linqi Ye, Rankun Li, Xiaowen Hu, Jiayi Li, Boyang Xing, Yan Peng, Bin Liang

*Abstract*—This paper introduces Unity RL Playground, an open-source reinforcement learning framework built on top of Unity ML-Agents. Unity RL Playground automates the process of training mobile robots to perform various locomotion tasks such as walking, running, and jumping in simulation, with the potential for seamless transfer to real hardware. Key features include one-click training for imported robot models, universal compatibility with diverse robot configurations, multi-mode motion learning capabilities, and extreme performance testing to aid in robot design optimization and morphological evolution. The attached video can be found at https://linqi-ye.github.io/video/iros25.mp4 and the code is coming soon.

## I. INTRODUCTION

Reinforcement learning (RL) has emerged as a powerful tool for training mobile robots to perform complex locomotion tasks [1-3]. However, the process of setting up simulation environments, defining reward functions, and importing new robots remains time-consuming and technically challenging. To address these issues, we present Unity RL Playground, a unified framework that significantly simplifies the RL workflow for mobile robots.

In recent years, the field of robotic reinforcement learning has witnessed remarkable advancements, thanks to the development of specialized frameworks that streamline the process of training and deploying RL algorithms on robotic systems. Below is a detailed overview of several prominent robotic RL frameworks:

Legged Gym [4] is a framework designed specifically for legged robots, facilitating the development of locomotion policies. By leveraging GPU-accelerated physics simulations, it enables researchers to train RL algorithms efficiently. The framework supports a range of legged robots and terrain types, providing a robust testbed for locomotion research. It emphasizes rapid experimentation and policy deployment, allowing researchers to bridge the sim2real gap more effectively.

Humanoid-Gym [5] is an open-source RL framework tailored for humanoid robots. Built upon NVIDIA Isaac Gym, it enables the training of locomotion skills with a focus on zero-shot sim2real transfer. Humanoid-Gym incorporates domain randomization and advanced reward functions to enhance the robustness of trained policies. The framework also supports a variety of humanoid robots, facilitating research into complex manipulation and locomotion tasks.

IsaacLab [6] is a comprehensive robotics simulation toolkit from NVIDIA. It provides high-fidelity physics and photorealistic rendering capabilities, enabling the creation of complex and realistic simulation environments. IsaacLab supports a wide range of robotic platforms and sensors, making it suitable for various robotics applications. The unified API allows for seamless integration of RL algorithms, making it an ideal platform for developing and testing advanced RL policies.

MuJoCo Playground [7] is an open-source framework for robot learning that leverages the MuJoCo physics engine. It is designed to facilitate rapid iteration and deployment of sim2real RL policies. The framework supports a variety of robotic platforms, including quadrupeds, humanoids, and dexterous hands. MuJoCo Playground incorporates on-device rendering and domain randomization techniques to enhance sim2real transfer. It also provides pre-built environments and benchmarks to streamline the development process.

Genesis [8] is a framework that focuses on generative physics modeling for robotics. It enables the creation of realistic and diverse simulation environments through a data-driven approach. Genesis supports various robotic tasks, including manipulation and locomotion. By leveraging generative models, it can generate new simulation scenarios on-the-fly, facilitating large-scale experimentation and policy training.

While these frameworks have significantly advanced the field of robotic RL, they come with certain limitations. The setup and configuration processes can be complex, requiring expertise in both robotics and RL. Development cycles are often lengthy, due to the need for extensive customization and tuning. Additionally, the high computational requirements, particularly the need for GPU acceleration, can limit the accessibility of these frameworks.

Unity ML-Agents [9] provides a user-friendly and versatile platform for RL research. Unity ML-Agents supports both Windows and Ubuntu operating systems and does not require GPU acceleration for training, making it accessible to a broader audience. Unity ML-Agents is designed to be intuitive and easy to use, with a focus on rapid development of games. However, it is not specifically tailored for robotic applications.

Building upon the strengths of Unity ML-Agents, we have developed Unity RL Playground, a dedicated RL framework for mobile robots. Unity RL Playground shares a unified codebase,

L. Ye, R. Li, X. Hu, Y. Peng are with the School of Future Technology, Shanghai University, 200444 Shanghai, China.
J. Li, B. Liang are with the Navigation and Control Research Center, Department of Automation, Tsinghua University, 100084 Beijing, China.
B. Xing is with the National and Local Co-Built Humanoid Robotics Innovation Center.

ensuring consistency across different robot platforms. It is designed to be operated with minimal programming expertise, allowing users to easily import their custom robot models for comprehensive multi-modal motion training.

The primary contribution of this work is the development of Unity RL Playground, a novel framework that greatly simplifies the complexity of robotic RL frameworks. By leveraging the strengths of Unity ML-Agents and tailoring it specifically for mobile robots, Unity RL Playground provides a "one-click" solution for training and deploying RL policies. Users can import various robot models into the simulation environment and, without any programming, enabling them to automatically learn multiple motion skills such as walking, running, and jumping. The framework also facilitates easy migration of trained policies to real hardware.

## II. UNITY RL PLAYGROUND OVERVIEW

### A. Software Interface

Unity RL Playground consists of a main menu interface and a training interface, as shown in Fig. 1 and Fig. 2, respectively.

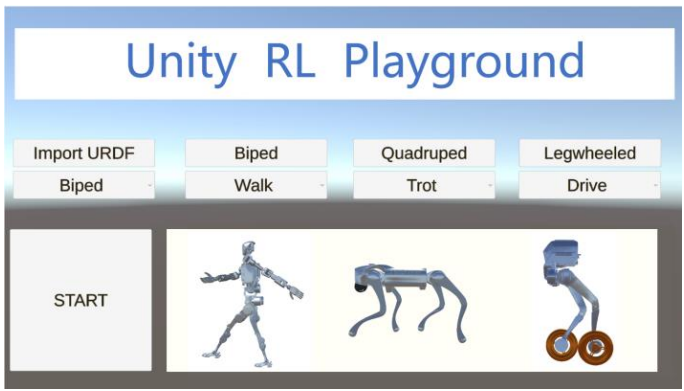

Figure 1. The main menu of Unity RL Playground. *The user can import their robot model and start training on one click.*

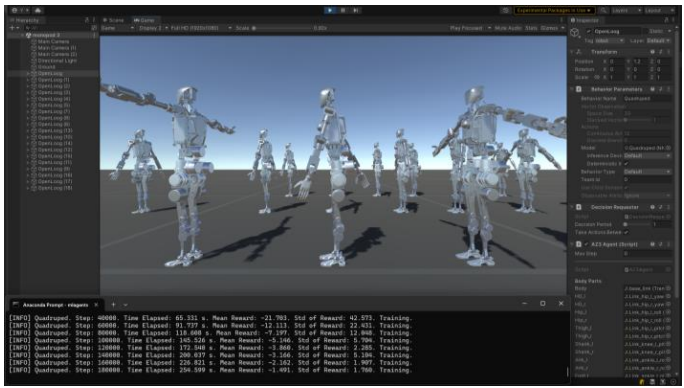

Figure 2. The training interface. *The robot is trained parallelly with CPU.*

### B. Key Features

(1) One-Click Training for Imported Robot Models

Users import a robot URDF file and select the training goal. The framework automatically generates locomotion policies by iteratively optimizing reward functions through reinforcement learning pipeline. It eliminates manual reward shaping and tedious coding, reducing setup time from days to minutes.

(2) Universal Configuration Compatibility

The framework dynamically adjusts simulation parameters to accommodate diverse robot morphologies, including bipeds, quadrupeds, and wheeled robots. It enables rapid prototyping by testing how structural variations (e.g., leg length, wheel diameter) affect motion performance, facilitating morphology-aware RL training.

(3) Multi-Mode Motion Learning

Unity RL Playground leverages the instruction learning technique [10] to learn various locomotion behaviors. The framework autonomously switches between walking, running, and jumping based on the selected task objective.

(4) Extreme Performance Testing

Unity RL Playground possesses the capability to undertake extreme performance testing by exploring the performance boundaries of robots under given constraints. This includes pushing robots to their limits in various scenarios, such as extreme terrains, high speeds, and heavy loads, to evaluate their durability, agility, and adaptability.

### C. Technical Highlights

The underlying learning framework of the Unity RL Playground is depicted in Fig. 3. This framework showcases two notable technical highlights: the Instruction Learning Technique and the Adaptive Learning Curriculum.

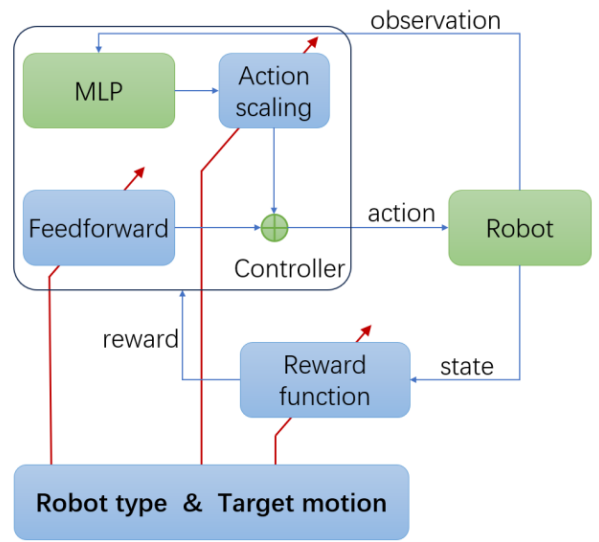

Figure 3. The learning framework of Unity RL Playground.

(1) The Instruction Learning Technique

The instruction learning technique, introduced in our prior work [10], leverages a combination of a feedforward signal and a scaled output from a neural network (MLP) as the action sent to the robot. This approach offers enhanced efficiency and flexibility compared to imitation learning. In the Unity RL Playground, an auto-configuration module automatically selects the feedforward signal, action scaling factor, and reward function weights based on the robot type and target motion.

(2) The Adaptive Learning Curriculum

We have further developed an innovative adaptive curriculum learning method that plays a pivotal role in optimizing the training process within the Unity RL Playground. This method meticulously adjusts the weights of the reward function in a dynamic and adaptive manner throughout the training period. By doing so, it ensures that the robotic agent can more effectively and efficiently learn to perform the desired target motion. This adaptive approach takes into account various factors, such as the progress of the training, the current performance of the agent, and the complexity of the target motion, to tailor the learning experience and enhance the overall outcome.

## III. FRAMEWORK ARCHITECTURE

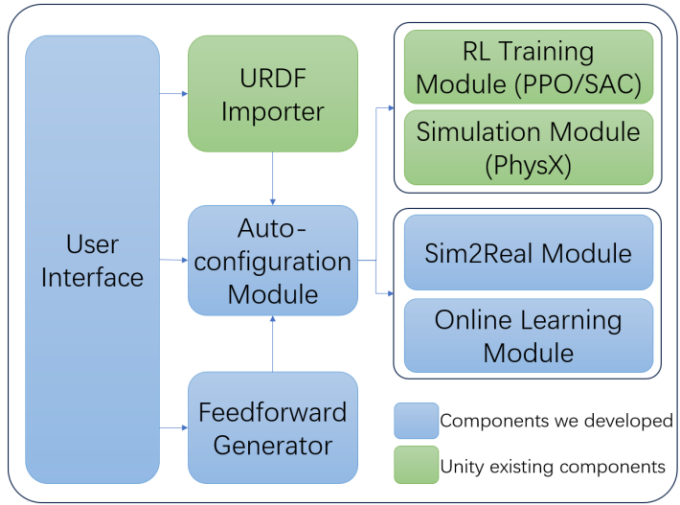

Figure 4. Components of Unity RL Playground.

### A. Components

The Unity RL Playground is composed of several key modules that work together to facilitate mobile robots RL tasks. We utilize both the components developed specifically for Unity RL Playground and existing Unity components, leveraging the extensive functionality and features provided by the Unity platform.

Starting with the user interface (see Fig. 1), which provides the user with an interactive environment to import robot model and select the target motion. The feedforward generator can generate a reference motion for the robot as a feedforward motor command. The URDF importer module is responsible for importing the robot URDF (United Robotics Description Format) files, essential for setting up robotic models. The auto-configuration module streamlines the process of configuring RL parameters and settings, ensuring that the training and simulation modules are correctly set up for the imported robot model. This module feeds into both the RL training module (supporting algorithms like PPO and SAC) and the simulation module (powered by PhysX), allowing for effective training and physical simulations of the robotic models. The sim2real module can send motor commands to the real robot through EtherNet for real-world implementations, helping to ensure that trained policies can be easily deployed in physical robots. Furthermore, we have developed a state alignment tool that enables real-time comparison of states between the real robot and the simulation model, facilitating the alignment of real robot states with those in simulation for swift migration of trained behaviors. The online learning module represents a novel endeavor, combining the sim2real communication and the RL training module to leverage real-world robotic motion data for training, thereby overcoming the sim2real discrepancy.

Overall, these modules work in unison to provide a comprehensive and efficient framework for RL research and development within Unity.

### B. Workflow

(1) Import robot URDF

The user clicks "Import URDF" to import their robot model. Then select the robot type (currently three options available: biped, quadruped, or leg-wheeled). The framework automatically configures the model for simulation.

(2) Select target motion

The user selects the simulation task objectives from the drop-down box. We provide three motion objectives for each kind of robot. They are: walk, run, and jump for biped robot; trot, bound, and pronk for quadruped robot; drive, walk, and jump for leg-wheeled robot. The framework uses the automatic configuration module to initialize the environment and prepare it for training.

(3) Start training

The user clicks "Start" button to start the RL training. The RL training module applies RL algorithms to optimize the robot's locomotion policy. Training progress is monitored in real-time through Unity's visualization tools.

Once trained, the system outputs the optimized locomotion policy. The user can visualize and test the generated motions within the simulation environment.

## IV. RESULTS

### A. Simulation Training

We have utilized the Unity RL Playground to train various models of bipedal, quadrupedal, and leg-wheeled robots on diverse tasks such as walking, running, and jumping. The simulation results have demonstrated their remarkable ability to swiftly learn and execute various movements. Furthermore, we have tested the performance of the bipedal robots in challenging environments, including rotating staircases laden with obstacles and realistic household settings. These tests have showcased their robustness and adaptability, further validating the effectiveness of our training approach. The robots have consistently demonstrated smooth and efficient movement patterns, even in complex and dynamic scenarios, underscoring their potential for real-world applications. Some of the results are shown in Figs. 5-9.

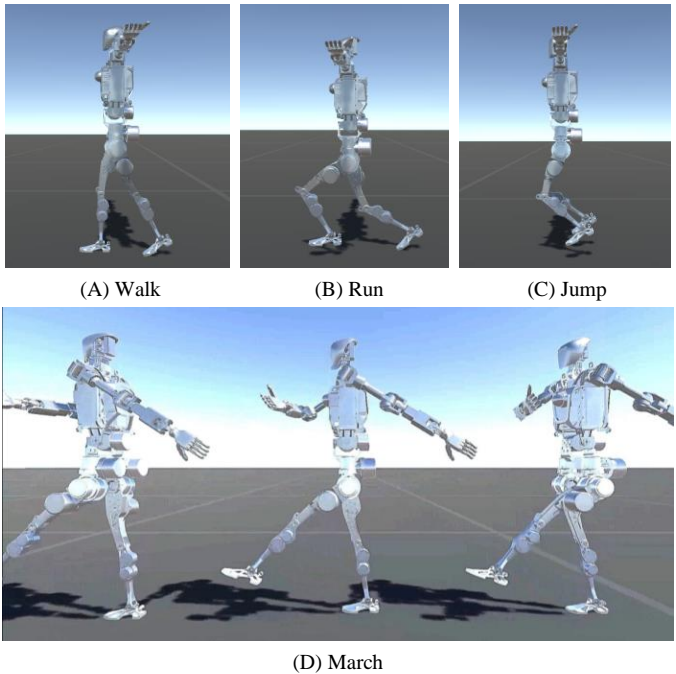

(A) Walk  (B) Run  (C) Jump

(D) March

Figure 5. The trained motion for a biped robot (Openloong).

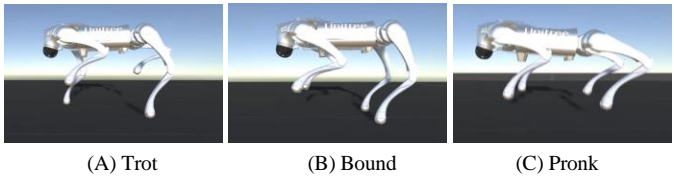

(A) Trot  (B) Bound  (C) Pronk

Figure 6. The trained motion for a quadruped robot (Unitree Go2).

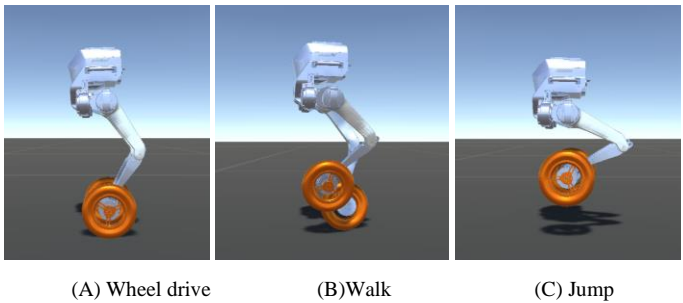

(A) Wheel drive  (B)Walk  (C) Jump

Figure 7. The trained motion for a leg-wheeled robot (Tron1).

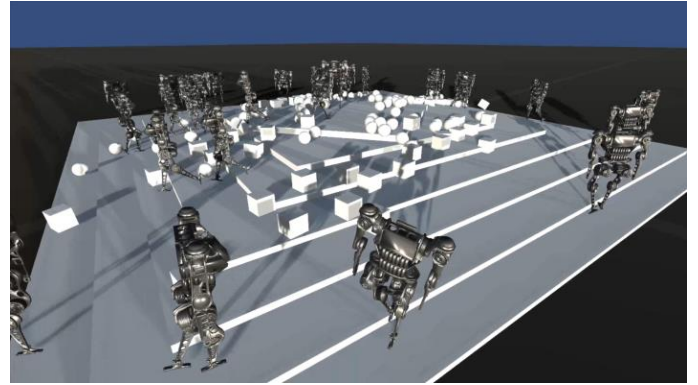

Figure 8. Biped robot tested on stairs with obstacles (Kuavo).

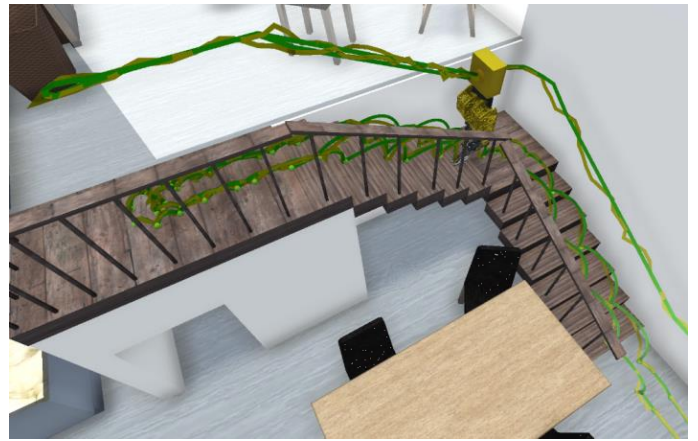

Figure 9. Biped robot tested in household scene (Ranger Max).

The Unity RL Playground stands out for its impressive versatility and extensive scalability. With minimal modifications, it can seamlessly adapt to accommodate a wide array of robotic types beyond the conventional ones, such as the monopod robot, jumping bicycle, point-feet biped, and straight-legged quadruped, among others, as shown in Fig. 10.

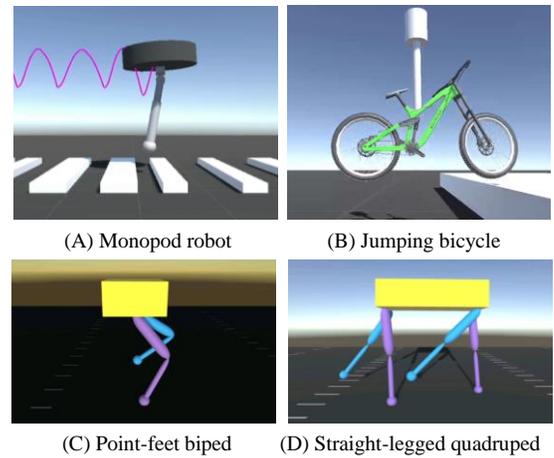

(A) Monopod robot  (B) Jumping bicycle

(C) Point-feet biped  (D) Straight-legged quadruped

Figure 10. Unity RL Playground extendibility.

## B. Sim2Real Experiments

We tested the trained policy for a biped robot and a quadruped robot using the sim2real module. The experiments involved transferring the learned behaviors from simulations to real-world scenarios, encompassing bipedal walking, quadrupedal walking, and quadrupedal jumping, as shown in Fig. 11 and Fig. 12. The results demonstrate the effectiveness of our trained policies in real-world conditions. The biped robot exhibited smooth and stable walking gaits, while the quadruped robot showed agile and coordinated movements, including proficient jumping abilities. These outcomes validate the robustness and adaptability of our sim2real approach, reinforcing the potential of Unity RL Playground for facilitating the seamless deployment of robot control policies in real-world applications.

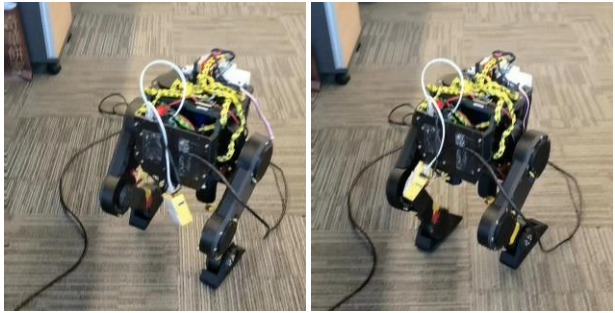

Figure 11. Sim2real experiment for a biped robot (Tinker).

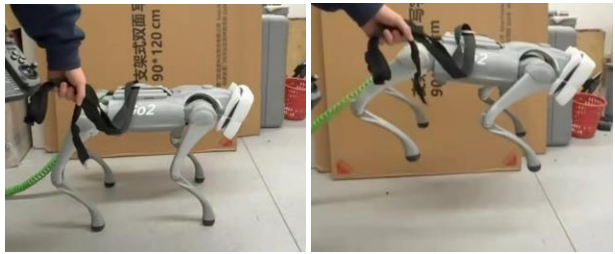

Figure 12. Sim2real experiment for a quadruped robot (Unitree Go2).

## C. Real-World Online Learning

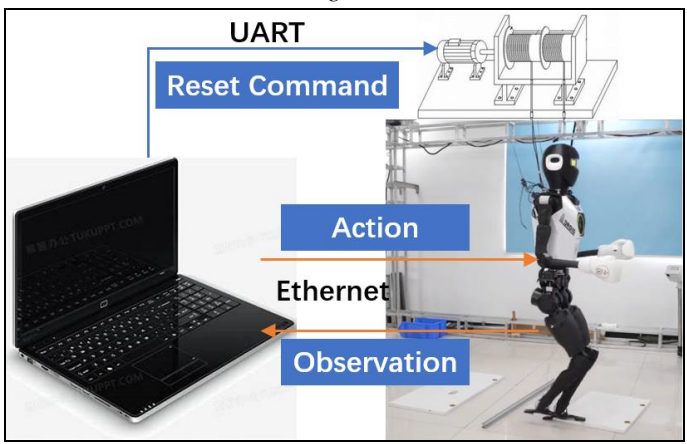

Figure 13. The real-world online learning system (X02-Lite).

In addition to facilitating the transfer from simulation to reality, Unity RL Playground can also be utilized for real-world online learning. We have developed an online learning system for biped robots based on Unity RL Playground, as illustrated in Fig. 13. This system features an automatic reset capability, enabling continuous learning and performance enhancement through minimal human intervention. By leveraging online learning, our system offers a novel pathway to bridge the sim2real gap for robots, paving the way for advanced locomotion behavior and enhanced adaptability in real-world environments.

## D. Robot Structure Optimization

Utilizing the versatility and extreme performance testing capabilities of Unity RL Playground, we can delve into identifying the optimal robot configurations under certain constraints. For example, Fig. 14 showcases an exploration of how leg length affects walking speed, while Fig. 15 presents an analysis of the impact of yaw joints on turning performance. These investigations highlight the potential of Unity RL Playground in facilitating the optimization of robotic designs tailored to specific operational requirements.

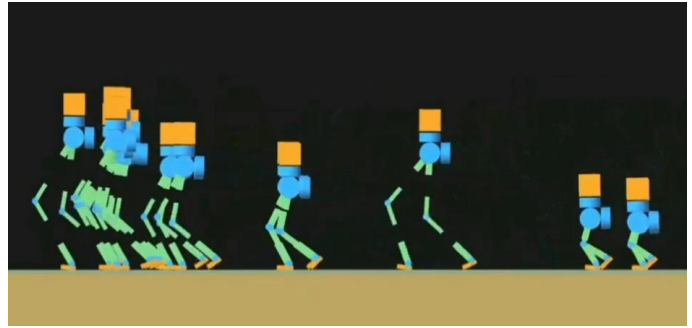

Figure 14. Robot performance test for leg length.

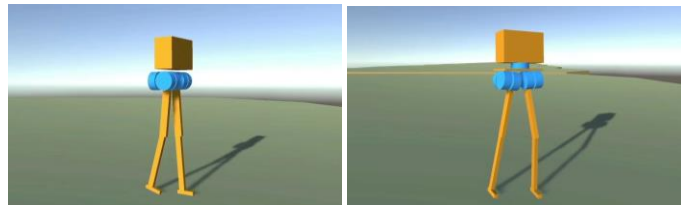

Figure 15. Robot performance test for hip yaw joint.

## V. DISCUSSION

Compared to other robot reinforcement learning frameworks like Legged Gym, IsaacLab, and Mujoco Playground, Unity RL Playground offers a distinctive set of advantages that make it stand out. Its one-click training feature significantly simplifies the process, lowering the barrier to entry for those new to RL-based robot training. Furthermore, its versatility allows for universal compatibility with a wide range of robot types and structures. Despite these strengths, there are still areas where Unity RL Playground can be improved, such as further validating its sim2real transfer performance and extending its capabilities to support more complex tasks beyond basic locomotion.

*A. Advantages*

- **Simplicity**: One-click training significantly lowers the barrier to entry for RL-based robot training.
- **Versatility**: Universal compatibility with diverse robot types and structures.
- **Efficiency**: Automates the RL workflow, reducing the time and effort required for coding.

*B. Limitations and Future Work*

- **Real-World Complexity**: Further testing is needed to fully validate the sim2real transfer performance.
- **Extension to More Complex Tasks**: We plan to extend the framework to support tasks beyond basic locomotion, e.g., whole-body control.
- **User Interface Improvements**: Enhancements to the user interface for better usability and accessibility.

## VI. CONCLUSION

Unity RL Playground is designed to streamline and simplify the complex process of training mobile robots using reinforcement learning, particularly when it comes to integrating new robots. Its distinctive features, including one-click training, universal compatibility, multi-mode motion capabilities, and extreme performance testing, position it as a highly versatile and powerful tool for robot locomotion control and design optimization. With ongoing development and further testing, Unity RL Playground holds the transformative potential to reshape the field of mobile robot locomotion. Our vision is to establish Unity RL Playground as a product akin to Arduino, thereby significantly lowering the barrier to entry for robot reinforcement learning development and broadening its appeal to a wider audience.